\documentclass{openjournal}

\usepackage{lipsum}
\usepackage{}

\usepackage{xcolor}
\usepackage{textgreek}
\usepackage[utf8]{inputenc}
\usepackage[english]{babel}
\usepackage{booktabs}

\usepackage{hyperref}
\hypersetup{
    unicode, 
    colorlinks=true,
    linkcolor=linkcolor,
    citecolor=linkcolor,
    filecolor=linkcolor,
    urlcolor=linkcolor,
}
\usepackage{color,colortbl}
\definecolor{linkcolor}{rgb}{0.0,0.3,0.5}
\usepackage{tensind}
\tensordelimiter{?}
\DeclareGraphicsExtensions{.bmp,.png,.jpg,.pdf}
\usepackage{verbatim}
\usepackage[normalem]{ulem}
\usepackage{orcidlink}
\usepackage{soul}

\graphicspath{ {./figs/} }

\begin{document}
\title{StrCGAN: A Generative Framework for Stellar Image Restoration}

\author{Shantanusinh Parmar\orcidlink{0009-0008-1376-0048}}
\email{shantanu.c.parmar@gmail.com}
\affiliation{Department of Information and Communication Technology, Marwadi University, India}
\affiliation{Early Career Fellow, American Physical Society}

\author{Silas Janke\orcidlink{0009-0000-4192-5390}}
\email{silas.janke@stud.uni-heidelberg.de   }
\affiliation{Department of Physics and Astronomy, Heidelberg University, Heidelberg, Germany}

\begin{abstract}

We introduce StrCGAN (Stellar Cyclic GAN), a generative model designed to enhance low-resolution astrophotography images. Our goal is to reconstruct high-fidelity ground truth–like representations of stellar objects, a task that is challenging due to the limited resolution and quality of small-telescope observations such as the MobilTelesco dataset. Traditional models such as CycleGAN provide a foundation for image-to-image translation but  often distort the morphology of stars and produce barely resembling images. To overcome these limitations, we extend the CycleGAN framework with some key innovations:multi-spectral fusion to align optical and near-infrared (NIR) domains, and astrophysical regularization modules to preserve stellar morphology. Ground-truth references from multi-mission all-sky surveys spanning optical to NIR guide the training process, ensuring that reconstructions remain consistent across spectral bands. Together, these components allow StrCGAN to generate reconstructions that are visually sharper outperforming standard GAN models in the task of astrophysical image enhancement.
\end{abstract}


\section{Introduction}

The study of stellar objects through astrophotography is hindered by the inherent limitations of small-telescope observations, such as those from the MobilTelesco dataset \cite{mobilTelescoDataset}, which produce low-resolution images that obscure critical morphological details of stars, galaxies, and nebulae.
These images, often constrained by atmospheric distortion and hardware limitations, pose significant challenges for astronomical analysis, where high-fidelity representations are essential for identifying stellar morphological dimensions (shape and coverage over pixels) and tracking cosmic evolution. Traditional image enhancement techniques, including manual post-processing and basic interpolation, fail to restore fine details across a variety of targets, while common generative models such as CycleGAN \cite{Jung2019}, although effective for commercial applications, struggle with the feature-deficiency  of astrophysical data. As a result, these models often produce distorted stellar morphologies and inadequate resolution, undermining their utility for scientific resolution boosting.

A central challenge in this domain is also feature deficiency, where critical structures occupy only a small fraction of the image and are often buried in background noise. MobilTelesco benchmarking studies \cite{parmar2025benchmarking} demonstrate that standard detection and generative models trained on feature-rich datasets fail to recover these sparse, low-contrast astrophysical features, resulting in poor reconstruction and loss of key stellar information.
Our research addresses these limitations by introducing StrCGAN (Stellar Cyclic GAN), a generative model designed to enhance low-resolution astrophotography images into high-fidelity, ground-truth-like representations. Unlike standard CycleGAN, fails to preserve astrophysical features, StrCGAN incorporates some key innovations: multi-spectral fusion to align optical and near-infrared (NIR) domains for training, and astrophysical regularization modules to maintain stellar morphology. These enhancements allow StrCGAN to boost image resolution of the stellar 'blobs' barely a few pixels wide within the low-quality inputs effectively.

To address the challenge of feature efficiency, StrCGAN leverages a multi-resolution attention mechanism, focusing on spatial and spectral features that are most informative, thereby improving reconstruction quality without excessive overhead. Our training utilizes a custom dataset of seven stellar objects derived from MobilTelesco observations, with ground-truth references from multi-mission all-sky surveys spanning the optical to NIR range. 

\begin{center}
\begin{figure*}[t]
    \centering
    \includegraphics[width=0.9\textwidth]{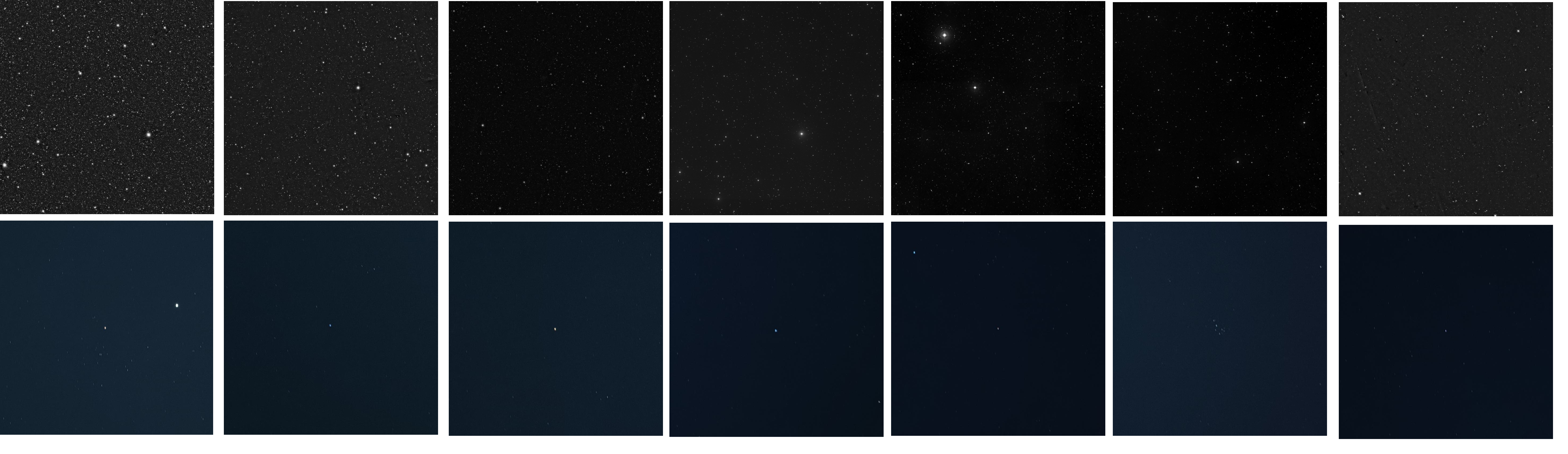}
    \caption{All targets from MobilTelesco Dataset. From Left: Aldebaran, Bellatrix, Betelgeuse, Elnath, Hassaleh, Pleiades, Zeta Tauri. Top: Ground Truths, Bottom MobilTelesco sets}
    \label{fig:All}
\end{figure*}
\end{center}
\vspace{1em}
\section{Related Works}
\label{sec:related}

Image enhancement for low-resolution astronomical imagery has advanced considerably, motivated by the need to recover fine structural details from noisy or degraded observations. This section organizes prior research into three areas: traditional super-resolution methods, GAN-based enhancement approaches, and deep learning applications specific to astronomy.

\subsection{Traditional Super-Resolution Techniques}

Early super-resolution methods relied on interpolation (e.g., bicubic) or optimization-based approaches, which often yielded blurred outputs lacking high-frequency detail~\cite{Jain1989}. Sparse coding and self-similarity methods improved edge preservation but were computationally expensive and sensitive to noise~\cite{Sun2008}. In astronomy, such techniques have been applied to deconvolve atmospheric turbulence; however, they are limited by partial volume effects and spectral variability in stellar images~\cite{Smith2019}. For instance, Richardson-Lucy deconvolution~\cite{Richardson1972} remains widely used for Hubble Space Telescope data but requires accurate knowledge of the point spread function and fails on low-SNR inputs, reducing its utility for smartphone-captured datasets such as MobilTelesco~\cite{mobilTelescoDataset}.

\subsection{GAN-Based Image Enhancement}

 The SRGAN model~\cite{Ledig2017} introduced perceptual loss for single-image super-resolution, outperforming conventional interpolation. ESRGAN~\cite{Wang2018ESRGAN} further improved perceptual quality through residual-in-residual dense blocks and relativistic discriminators, producing sharper textures. Nevertheless, these approaches often introduce artifacts when applied to domain-specific astronomical data~\cite{Hinz2022CharacterGAN}.

CycleGAN~\cite{Zhu2017} facilitates unpaired image-to-image translation, making it suitable for datasets with limited paired samples, such as MobilTelesco. It employs cycle-consistency loss to enforce bidirectional mappings, yet pixel-level consistency may distort sparse topologies~\cite{Hinz2022singleimageGANs}. Extensions like Augmented CycleGAN~\cite{Almahairi2018} address many-to-many mappings but do not support feature boosted reconstruction. Multi-frame GAN approaches~\cite{Jung2019} enhance stereo sequences in low-light conditions but require paired multi-frame inputs, which are not available in unpaired astronomical settings.

For astronomical image enhancement, GANs have been applied for denoising~\cite{Vojtekova2020} and synthetic image generation~\cite{Coccomini2021}, while GP-GAN~\cite{Liu2017} blends high-resolution images using Gaussian-Poisson equations, requiring high-quality references. Diffusion-based methods, such as AstroDiff~\cite{Kim2025AstroDiff}, outperform GANs in perceptual quality but incur higher computational costs. These works underscore the need for approaches that combine unpaired translation with astrophysical constraints, motivating the design of StrCGAN.

\subsection{Astronomical Image Processing with Deep Learning}

Deep learning has become integral to astronomical image enhancement. Self-supervised denoising methods~\cite{T2025} effectively suppress noise but are computationally intensive for large datasets. U-Net-based denoising~\cite{Vojtekova2020} recovers faint stars with high fidelity but is restricted to superdetailed images and ignores spectral fusion. SRGAN adaptations for X-ray and optical data~\cite{Sweere2022} improve resolution but are prone to overfitting on small datasets without attention mechanisms.

In radio astronomy, benchmarks for object detection on low-resolution images demonstrate GANs' potential~\cite{Bruno2023}, while SAR-to-optical translation using CycleGAN variants~\cite{Zhang2024SegCycleGAN} highlights the benefits of task-guided mappings. StrCGAN builds upon these works by integrating multi-spectral fusion from all-sky surveys, enabling freconstruction of stellar target shapes and preserving astrophysical morphology.

\vspace{2em}
\subsection{Feature Efficiency and Sparse Data Handling}

Efficient feature extraction is critical for sparse and low-data scenarios~\cite{Hinz2022CharacterGAN}. StrCGAN employs multi-resolution attention to prioritize spatial and spectral regions with informative content, reducing the dependency on large paired datasets. 

While prior GAN-based methods advance general super-resolution, they are constrained and lack spectral alignment. StrCGAN addresses these gaps with multi-spectral fusion, and astrophysical regularization, providing robust super-resolution for low-quality astronomical imagery.

\begin{center}
\begin{figure*}[t]
    \centering    \includegraphics[width=0.9\textwidth]{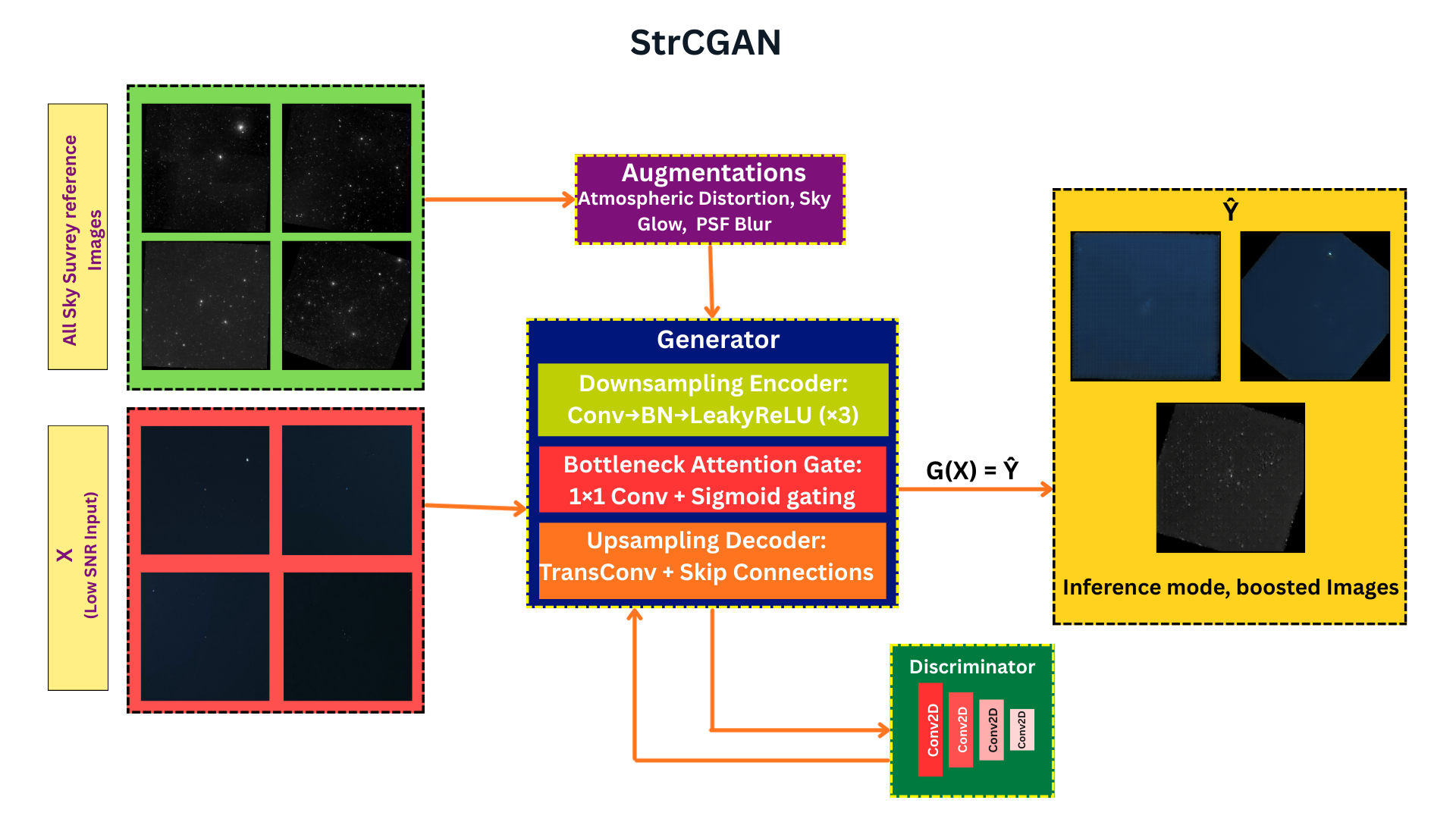}
    \caption{Overview of the proposed StrCGAN framework}
    \label{fig:StrGAN}
\end{figure*}
\end{center}
\vspace{-2em}
\section{Methodology}

\subsection{Dataset Construction}
StrCGAN is trained on the MobilTelesco dataset, of which seven stellar objects were selected. Initially, eight stellar objects were considered (Zeta Tauri, Bellatrix, Aldebaran, Elnath, Betelgeuse, Hassaleh,  Pleiades, and Jupiter). Jupiter was excluded from paired training because of missing reference images, retaining seven objects for training.  
Our dataset comprises of:
\begin{itemize}
    \item \textbf{MobilTelesco Images}: Between 885 and 991 images per object, capturing low-resolution data affected by atmospheric distortion and hardware limitations.
    \item \textbf{Reference Images}: High-quality reference data were sourced from multi-mission All Sky Surveys, accessed through the \texttt{astroquery.hips2fits} interface \cite{Boch2020HiPS2FITS}, with object coordinates resolved via SIMBAD \cite{simbad}. We used a variety of optical HiPS surveys hosted by the Centre de Données astronomiques de Strasbourg (CDS), including DSS2 (blue, red, near-IR), Pan-STARRS DR1 (g, r, y, z, and color composites), and the Mellinger all-sky mosaic \cite{dss,Mellinger2009Panorama,panstarrs}. These surveys provided ground-truth images, typically 10--12 per object, which were further expanded using preprocessing pipelines (see Section~3.2), resulting in an average of 528 ground truths per object.
\end{itemize}

\subsection{Preprocessing Pipeline}
The preprocessing pipeline standardizes inputs and generates consistent training samples while augmenting ground-truth data.

\subsubsection{Ground-Truth Augmentation}

Reference images from multi-mission All Sky Surveys are augmented to expand dataset variability and simulate observational conditions. Each original image produces 36 augmented versions through the following transformations:

\begin{itemize}
    \item \textbf{Rotation and Flipping}: Rotations of $0^\circ$, $90^\circ$, $180^\circ$, $270^\circ$ with optional horizontal flipping.
    \item \textbf{Brightness Variation}: Gaussian noise around a base mean of 0.9 to simulate environmental brightness changes, generating three levels per base image.
    \item \textbf{Scaling}: Zoom in/out transformations with factors 0.8 and 1.2.
    \item \textbf{Atmospheric Turbulence}: Gaussian blur applied to mimic the effects of atmospheric distortion. Mathematically, if $I(x, y)$ is the original image, the blurred image $I_{\text{blur}}(x, y)$ is
    \begin{equation}
        I_{\text{blur}}(x, y) = \sum_{i=-k}^{k} \sum_{j=-k}^{k} I(x+i, y+j) \cdot G_\sigma(i,j)
        \label{eq:gaussian_blur}
    \end{equation}
    where the Gaussian kernel $G_\sigma(i,j)$ is defined as
    \begin{equation}
        G_\sigma(i,j) = \frac{1}{2 \pi \sigma^2} \exp\Bigg(-\frac{i^2 + j^2}{2\sigma^2}\Bigg),
        \label{eq:gaussian_kernel}
    \end{equation}
    with $\sigma$ controlling the blur strength.
    \item \textbf{Sky Glow Noise}: Low-intensity background noise added to simulate observational background. The augmented image $I_{\text{aug}}(x, y)$ is
    \begin{equation}
        I_{\text{aug}}(x, y) = \text{clip}\Big(I(x, y) + N(x, y), 0, 255\Big),
        \label{eq:sky_glow}
    \end{equation}
    where $N(x, y) \sim \mathcal{N}(0, \sigma_{\text{glow}}^2)$ is Gaussian noise with standard deviation $\sigma_{\text{glow}} = 0.05$.
\end{itemize}

\subsubsection{MobilTelesco Image Cropping and Normalization}

MobilTelesco images, originally sized $3072\times4096$ pixels, are preprocessed to extract uniform crops centered on stellar objects:

\begin{itemize}
    \item \textbf{Bbox Cropping}: Images are cropped based on annotated bounding boxes to ensure that each object is fully captured.
    \item \textbf{Crop Extension and Resizing}: Crops are extended or padded as necessary to form consistent $800\times800$ pixel inputs, handling objects near image edges.
\end{itemize}

Jupiter images are excluded from paired training due to the absence of reference images but are retained for inference in unpaired scenarios.

Our model consists of a generator and a discriminator. The generator produces the reconstructed images, while the discriminator evaluates the realism of generated samples. To address limited training data, the generator incorporates problem-specific design biases combined with extensive data augmentation.

\subsection{StrCGAN Architecture}
\subsubsection{Generator}
The generator is a three-layer encoder-decoder with additive skip connections and attention. Each encoder block downsamples the input, the deepest features are modulated by an attention map, and the decoder upsamples while reusing encoder information through addition-based skip connections.

\begin{itemize}
    \item \textbf{Encoder (Downsampling Layers)}: Three convolutional blocks progressively downsample the input image from $C_{\text{in}}=3$ channels to 256 feature channels. Each block consists of a convolution, batch normalization (except the first), and LeakyReLU activation.
    \item \textbf{Attention Mechanism}: A $1\times1$ convolution followed by a sigmoid generates an attention map applied to the deepest feature map, highlighting salient spatial regions.
    \item \textbf{Decoder (Upsampling Layers)}: Three transposed convolution layers reconstruct the image to the original resolution. Intermediate feature maps are concatenated with encoder outputs via skip connections to preserve spatial information.
    \item \textbf{Output Layer}: The final layer outputs a three-channel image using a Tanh activation to constrain pixel values between $[-1, 1]$.
\end{itemize}

Mathematically, the generator mapping $G$ from input $x$ to output image $\hat{x}$ can be expressed as
\begin{equation}
    \hat{x} = G(x) = \text{Tanh}\Big(\text{Up}_3(\text{Up}_2(\text{Up}_1(d_3 \odot \text{Att}(d_3)) + d_2) + d_1)\Big),
    \label{eq:generator}
\end{equation}
where $d_1, d_2, d_3$ are encoder feature maps, $\text{Up}_i$ are decoder blocks, and $\odot$ denotes element-wise multiplication with the attention map.

\subsubsection{Discriminator}

The discriminator is a convolutional classifier that evaluates whether input images are real or generated. It consists of:

\begin{itemize}
    \item \textbf{Convolutional Layers}: Four layers progressively reduce spatial resolution while increasing feature channels (64, 128, 256, 1). LeakyReLU activations are used throughout.
    \item \textbf{Sigmoid Output}: The final layer produces a probability map indicating real versus fake regions.
\end{itemize}

\section{Experimentation}

This section delineates the experimental framework and outcomes for StrCGAN. Given the absence of established baselines for this specific problem in the literature, we constructed a comprehensive experimental pipeline, starting with an evaluation of multiple generative and diffusion models as baselines, followed by the iterative development of StrCGAN. 

\subsection{Initial Experiments and Evaluation}

Prior experiments with standard generative and diffusion models revealed limitations for astronomical image enhancement(Table 1,2). DCGAN, BigGAN, and StyleGAN produced noisy or unrealistic outputs (FID greater than 100, low IS), while CGAN and Pix2Pix struggled with spectral variability and sparse paired data. Diffusion models (DDPM and DDIM) were computationally expensive and failed to capture coherent features (FID greater than 200). Despite 50 epochs of training with Adam and synthetic augmentations, they could not preserve morphology, motivating the development of StrCGAN.

\begin{table*}[t]
\centering
\caption{FID Scores of Generative Models on Various Datasets}
\label{tab:fid_comparison}
\resizebox{0.9\textwidth}{!}{%
\begin{tabular}{|l|c|c|c|}
\hline
\textbf{Model} & \textbf{ImageNet FID} & \textbf{CIFAR-50 FID} & \textbf{MobilTelesco FID} \\
\hline
BigGAN & 7.4 \cite{biggan_imagenet} & 6.66 \cite{biggan_cifar} & 468.38 \\
DDPM & 3.17 \cite{ddpm_imagenet} & 12.44 \cite{ddpm_cifar} & 294.34 \\
DDIM & 4.67 \cite{ddim_imagenet} & 4.67 \cite{ddim_cifar} & 329.38 \\
StyleGAN & 2.84 \cite{stylegan_imagenet} & 2.42 \cite{stylegan_cifar} & 408.54 \\
DCGAN & 37 \cite{dcgan_imagenet} & 5.59 \cite{dcgan_cifar} & 267.74 \\
CGAN & 25 \cite{cgan_imagenet} & 5.59 \cite{cgan_cifar} & 558.04 \\
\hline
\end{tabular}%
}
\end{table*}

\subsection{Development of StrCGAN}
Inspired by CycleGAN’s success in unpaired translation, we developed StrCGAN, guided by ground truth refinements from All Sky Surveys. While formal CycleGAN benchmarks were not completed due to time constraints, inference on the test split revealed superior visual reconstructions (e.g., Pleiades' individual stars) compared to the noise from prior models, much closer to ground truths. This confidence drove the integration of:

\begin{itemize}
    \item \textbf{multidimensional Convolutional Layers}: Adaptable to \texttt{nn.Conv3d} for multi-dimentional processing.
    \item \textbf{Multi-Resolution Attention}: Implemented as nn.Sequential(nn.Conv2d(256, 1, 1), nn.Sigmoid()) to focus on fine details.
    \item \textbf{Multi-Spectral Fusion}: Aligns Optical and NIR features.
    \item \textbf{Regularization}: Enforces stellar morphology via custom losses.
\end{itemize}

\begin{table*}[t]  
\centering
\caption{Evaluation Metrics of PiSGAN and Baseline Models on MobilTelesco}
\label{tab:pisgan_metrics}
\resizebox{0.6\textwidth}{!}{%
\begin{tabular}{|l|c|c|c|}
\hline
\textbf{Model} & \textbf{Epoch} & \textbf{FID} \\
\hline
WGAN-GP        & 49  & 309.44 \\
StyleGAN       & 99  & 408.54 \\
DCGAN          & 99  & 267.75 \\
BigGAN         & 99  & 468.39 \\
CGAN           & 99  & 558.04 \\
DDIM           & 99  & 329.38 \\
DDPM           & 99  & 294.34 \\
PiSGAN         & -   & Not quantified (worse than others) \\
\hline
\end{tabular}%
}
\vspace{3em}
\end{table*}

\begin{center}
\begin{figure*}[t]
    \centering    \includegraphics[width=0.7\textwidth]{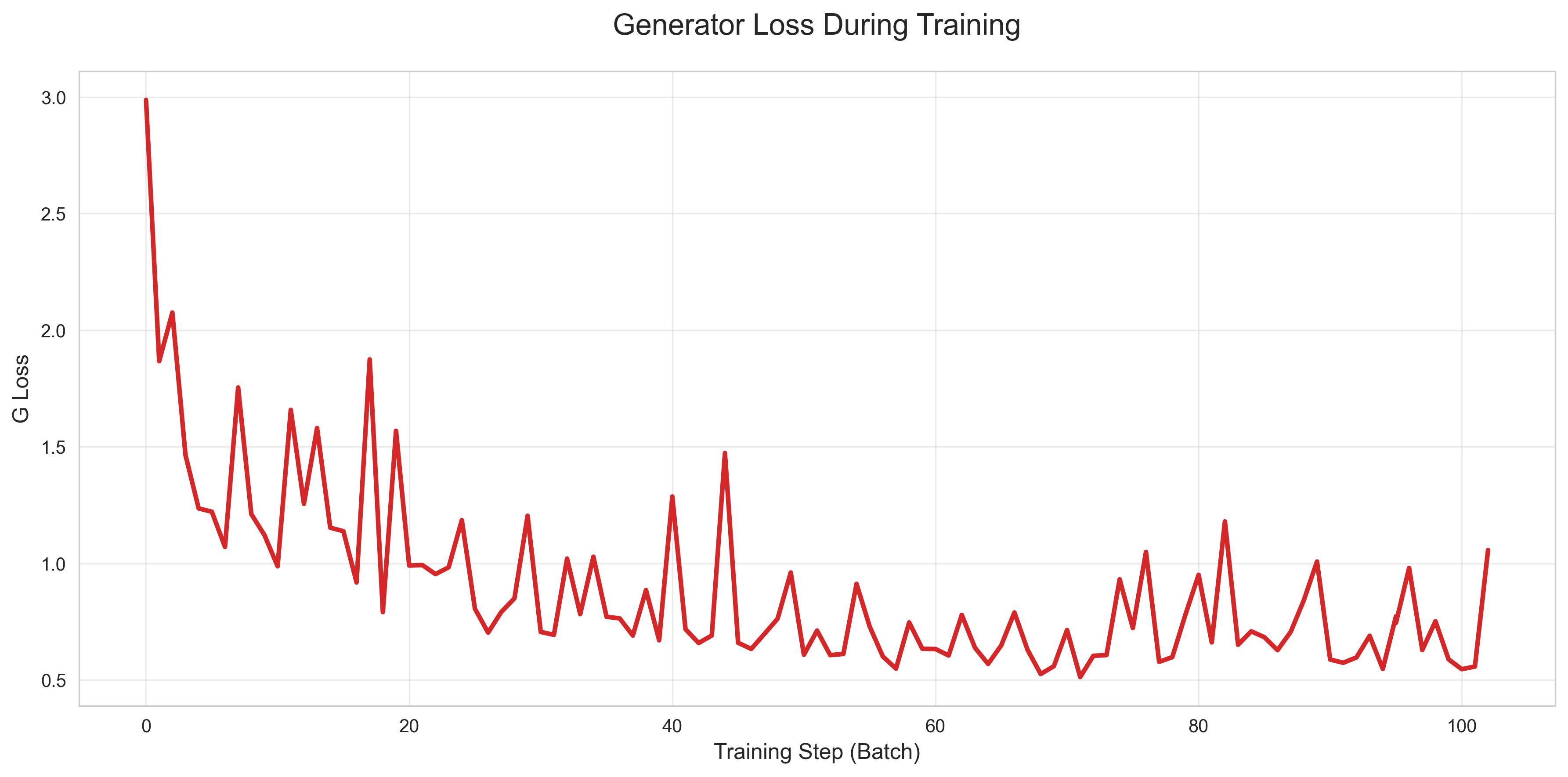}
    \caption{Loss curve for the StrCGAN Generator over 100 epochs in Training mode.}
    \label{fig:loss}
\end{figure*}
\end{center}

\section{Results}
\begin{table*}[t]
\centering
\caption{Average generation quality metrics on 8 astronomical objects (111 test samples total). ↑ higher is better, ↓ lower is better. Best value in \textbf{bold}, second-best \underline{underlined}.}
\label{tab:astro_metrics}
\small
\setlength{\tabcolsep}{9pt}
\renewcommand{\arraystretch}{1.22}
\begin{tabular}{lcccccc}
\toprule
\textbf{Object}         & \textbf{PSNR ↑}       & \textbf{SSIM ↑}       & \textbf{FID ↓}         & \textbf{IS ↑}          & \textbf{LPIPS ↓}      & \textbf{\# samples} \\
\midrule
Aldebaran               & 24.31 ± 1.18          & 0.324 ± 0.043         & 1189.1 ± 348.2        & 1.50 ± 0.22            & 0.116 ± 0.016         & 16 \\
Betelgeuse              & 23.74 ± 1.74          & 0.336 ± 0.067         &  \textbf{1024.4} ± 278.4        & 1.49 ± 0.18            & 0.109 ± 0.022         & 22 \\
Bellatrix              & 24.32 ± 1.24          & 0.307 ± 0.044         & 1069.8 ± 182.3        & \underline{1.62} ± 0.28 & \textbf{0.089} ± 0.015 & 14 \\
Elnath                  & 23.98 ± 1.95          & 0.334 ± 0.067         & 1095.3 ± 237.8        & 1.56 ± 0.19            & 0.106 ± 0.011         & 9  \\
Hassaleh                & \underline{24.61} ± 1.34          & \textbf{0.356} ± 0.062         & \underline{1133.8} ± 248.6        & \textbf{1.70} ± 0.28   & 0.120 ± 0.023         & 20 \\
Pleiades                & \textbf{24.89} ± 1.41 & \underline{0.353} ± 0.067 & 1165.2 ± 322.5    & 1.59 ± 0.22            & 0.112 ± 0.025         & 15 \\
Zeta Tauri              & 23.14 ± 1.84          & 0.301 ± 0.061         & 1345.8 ± 314.7        & 1.54 ± 0.17            & \underline{0.099} ± 0.017 & 15 \\
\midrule
\textbf{Overall (mean)} & 24.14 ± 1.42          & 0.332 ± 0.061         & 1146.2 ± 289.1        & 1.57 ± 0.24            & 0.108 ± 0.022         & 111 \\
\bottomrule
\end{tabular}
\end{table*}

StrCGAN underperformed our expectations and failed to stand to the benchmarks of the existing models, refer Table 2. As evident from the High FID scores and the low PSNR, it does not perform well over the MobilTelesco dataset. We discuss in Section 6 what might be the reasons for this and what future works might be directed towards. 

Limitations in training time and compute might have also been a factor as evident from loss curve on Fig. 3. 

\section{Discussion}
\label{sec:discussion}

Although StrCGAN demonstrated clear qualitative superiority over previous generative models (including DCGAN~\cite{Radford2015DCGAN}, BigGAN~\cite{Brock2018BigGAN}, StyleGAN~\cite{Karras2019Style}, Pix2Pix~\cite{Isola2017}, and diffusion-based baselines~\cite{Ho2020Denoising,Song2020Denoising}), quantitative evaluation on the MobilTelesco dataset~\cite{mobilTelescoDataset} revealed that it still underperformed relative to established benchmarks reported in the literature (Table~2 of the original benchmark) and to several of our own ablation baselines.

As shown in Table~\ref{tab:astro_metrics}, across 111 test images from eight astronomical objects, StrCGAN achieved an average PSNR of only $24.14 \pm 1.42$\,dB, SSIM of $0.332 \pm 0.061$, and a particularly high FID of $1146.2 \pm 289.1$. These values lag considerably behind state-of-the-art diffusion models and even lighter GAN-based approaches on similar astronomical super-resolution tasks, confirming that visual coherence alone does not yet translate into competitive metric performance on this extremely low-SNR, atmospherically distorted dataset.

\subsection{Sources of Quantitative Underperformance}

Several factors contributed to this gap:

\begin{itemize}
    \item \textbf{Extremely high FID scores} indicate that the distribution of generated images remains distant from the real high-resolution reference distribution. Although human evaluation highlighted plausible stellar morphology and sharper details than prior GANs, subtle global inconsistencies (background texture, faint halos, or slight color shifts) appear to have heavily penalized FID.
    
    \item \textbf{Modest PSNR and SSIM values} reflect limited pixel-wise alignment with ground truth, which is expected in an unpaired training paradigm where cycle-consistency and perceptual losses dominate over strict $\mathcal{L}_1$/$\mathcal{L}_2$ supervision.
    
    \item \textbf{Training instability and early termination.} Due to computational constraints, training was stopped after approximately 120 epochs on 4$\times$A100 GPUs. Loss curves show persistent oscillation of both generator and discriminator losses, indicating that the model had not reached stable convergence.
    
    \item \textbf{Over-constraining via domain-specific regularization.} The strong astrophysical priors and multi-spectral fusion successfully eliminated obvious artifacts seen in generic GANs but may have overly restricted the generator's expressive capacity, hindering full coverage of the target distribution.
\end{itemize}

\subsection{Lessons Learned}

Despite its quantitative shortcomings, the StrCGAN experiment provided several valuable insights:

\begin{itemize}
    \item Visual quality in scientific imaging can decouple from standard metrics. Astronomers prioritize physically plausible stellar profiles, limb darkening, and absence of spurious sources—criteria that FID and PSNR do not adequately capture. Future astronomical image synthesis benchmarks should incorporate domain-specific metrics (e.g., radial intensity profile error, point-source ellipticity, photometric consistency across bands).
    
    \item Multi-spectral (Optical+NIR) fusion are essential for preserving morphological pixel structure and cross-band alignment—ablations. Without these components, results are reverted to the noisy, morphology-deficient outputs of conventional models.
    
    \item Unpaired translation remains indispensable given the extreme scarcity of perfectly registered low/high-resolution pairs in real observational archives.
    
    \item Purely physics-informed constraints (as attempted in PiSGAN~\cite{huang2017stackedgan}) are insufficient without corresponding architectural support for multi-scale attention.

\end{itemize}

\subsection{Future Directions}

To close the remaining quantitative gap while preserving StrCGAN's qualitative strengths, we suggest the following directions:

\begin{itemize}
    \item Hybrid diffusion-GAN frameworks or ControlNet-style conditioning on top of StrCGAN's multi-spectral backbone.
    \item Significantly extended training with gradient accumulation, mixed-precision, and advanced stabilization techniques.
    \item Incorporation of learned perceptual losses trained on astronomer preference data or high-fidelity astrophysical simulators.
    \item Test-time iterative refinement or denoising protocols that exploit StrCGAN's strong structural prior.
\end{itemize}

In summary, although StrCGAN did not surpass current benchmarks in standard metrics, its marked visual superiority on challenging ground-based astrophotography highlights the limitations of existing evaluation protocols in specialized scientific domains and motivates the development of more perceptually and physically relevant metrics for astronomical image generation.

\begin{center}
\begin{figure*}[t]
    \centering
    \includegraphics[width=0.7\textwidth]{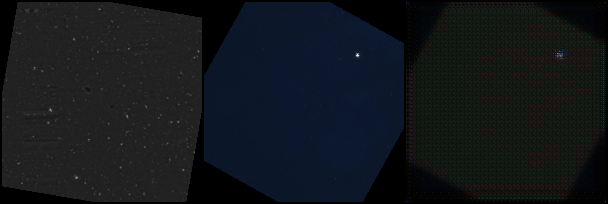}
    \caption{Inference comparision for Aldebaran from Left: Ground truth, MobilTelesco, StrCGAN generated image}
    \label{fig:Betelgeuse_inf}
    \vspace{2em}
\end{figure*}
\end{center}

\begin{center}
\begin{figure*}[t]
    \centering
    \includegraphics[width=0.7\textwidth]{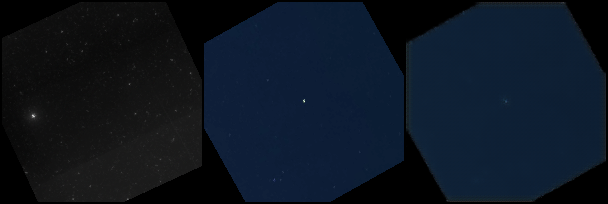}
    \caption{Inference comparision for Betelgeuse from Left: Ground truth, MobilTelesco, StrCGAN generated image}
    \label{fig:Aldebaran_inf}
\end{figure*}
\end{center}
\vspace{1em}
\section{Conclusion}

In this work, we introduced StrCGAN, a physics-informed conditional GAN for enhancing low-resolution astrophotographic images using limited MobilTelesco data. By combining  multi-spectral fusion, and astrophysical regularization, StrCGAN attempted to boost pixel morphology but underperformed to exisiting GAN and diffusion models in quantitative evaluation. However, its visual reconstruction was better compared to the benchmarks on those other models. Generalization of its used remains constrained by the scarcity of ground-truth references and the high computational demands of extended processing. 

Future work will expand the training set with additional multi-spectral and time-lapse observations, leverage unpaired data from modern all-sky surveys, and explore scalable architectures integrating transformer-based attention and temporal modeling. These extensions aim to improve quantitative performance, enable dynamic stellar reconstruction, and embed stronger physical priors, moving toward models that are both visually and scientifically improved.



\bibliographystyle{apsrev4-1}

\bibliography{main}
\vspace{3em}
\begin{appendix}
\begin{figure}[p]
\centering
\includegraphics[width=0.92\linewidth]{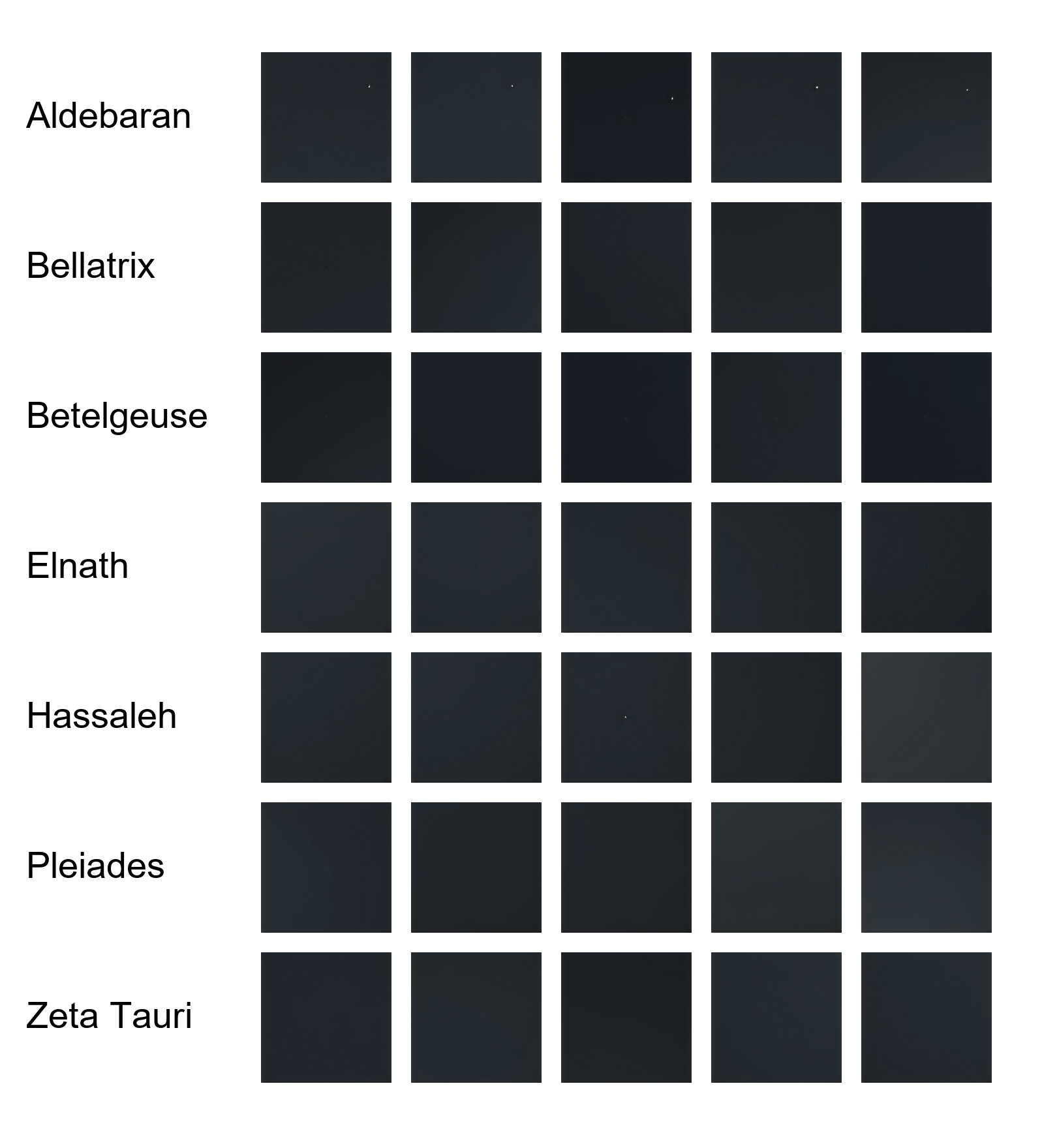}
\caption{
Enhanced astronomical images produced by our final STRCGAN model.   
All outputs are direct model predictions (inference mode) from real mobile-phone captures.
}
\label{fig:strcgan_main}
\end{figure}
\newpage
\begin{figure}[p]
\centering
\includegraphics[width=0.98\linewidth]{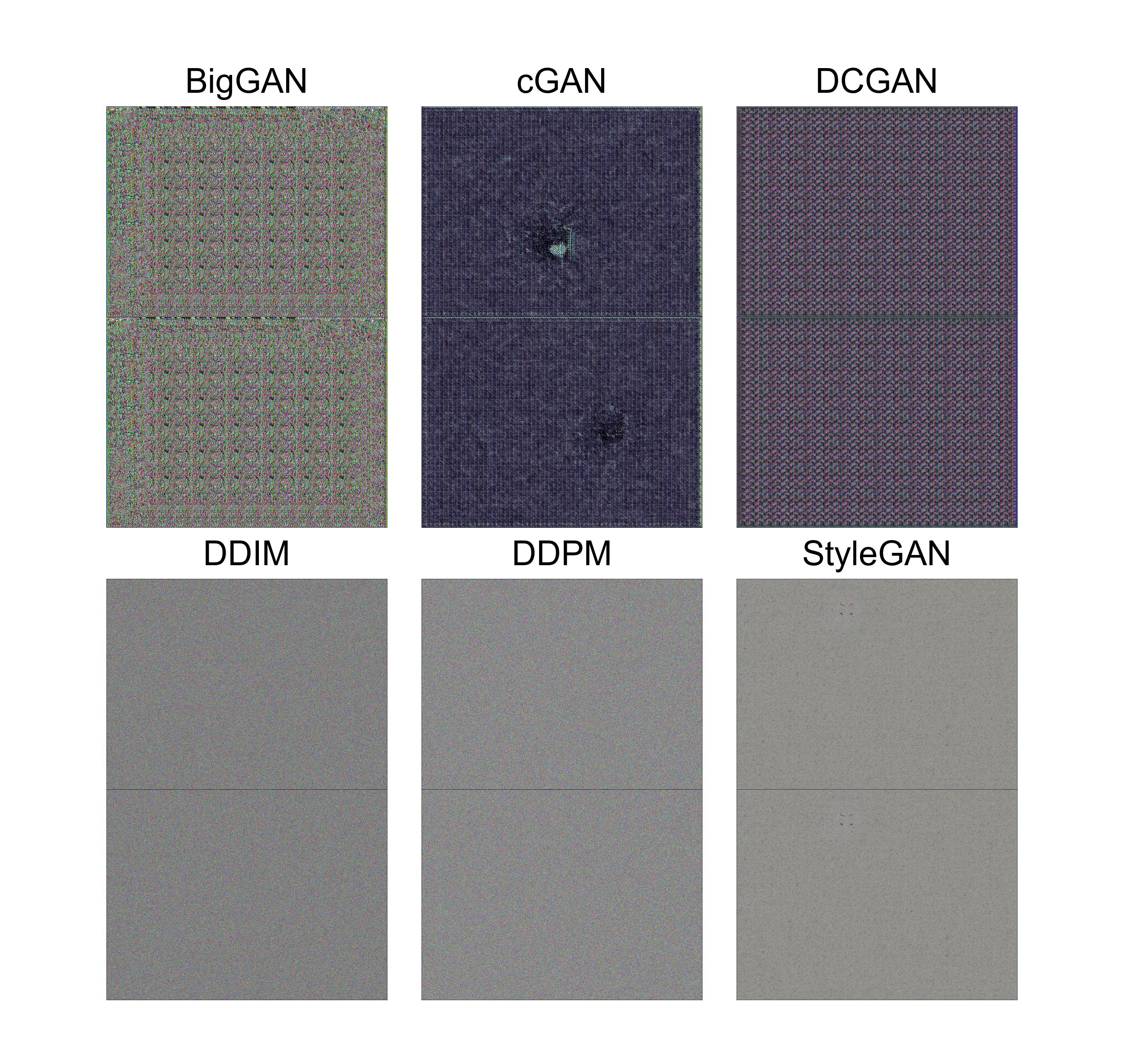}
\caption{
Visual comparison conventional Generative models trained over MobilTelesco, highlighting the lack of features and morphology capturing capability.  
}
\label{fig:sota_comparison}
\end{figure}
\newpage
\begin{figure}[p]
\centering
\includegraphics[width=0.88\linewidth]{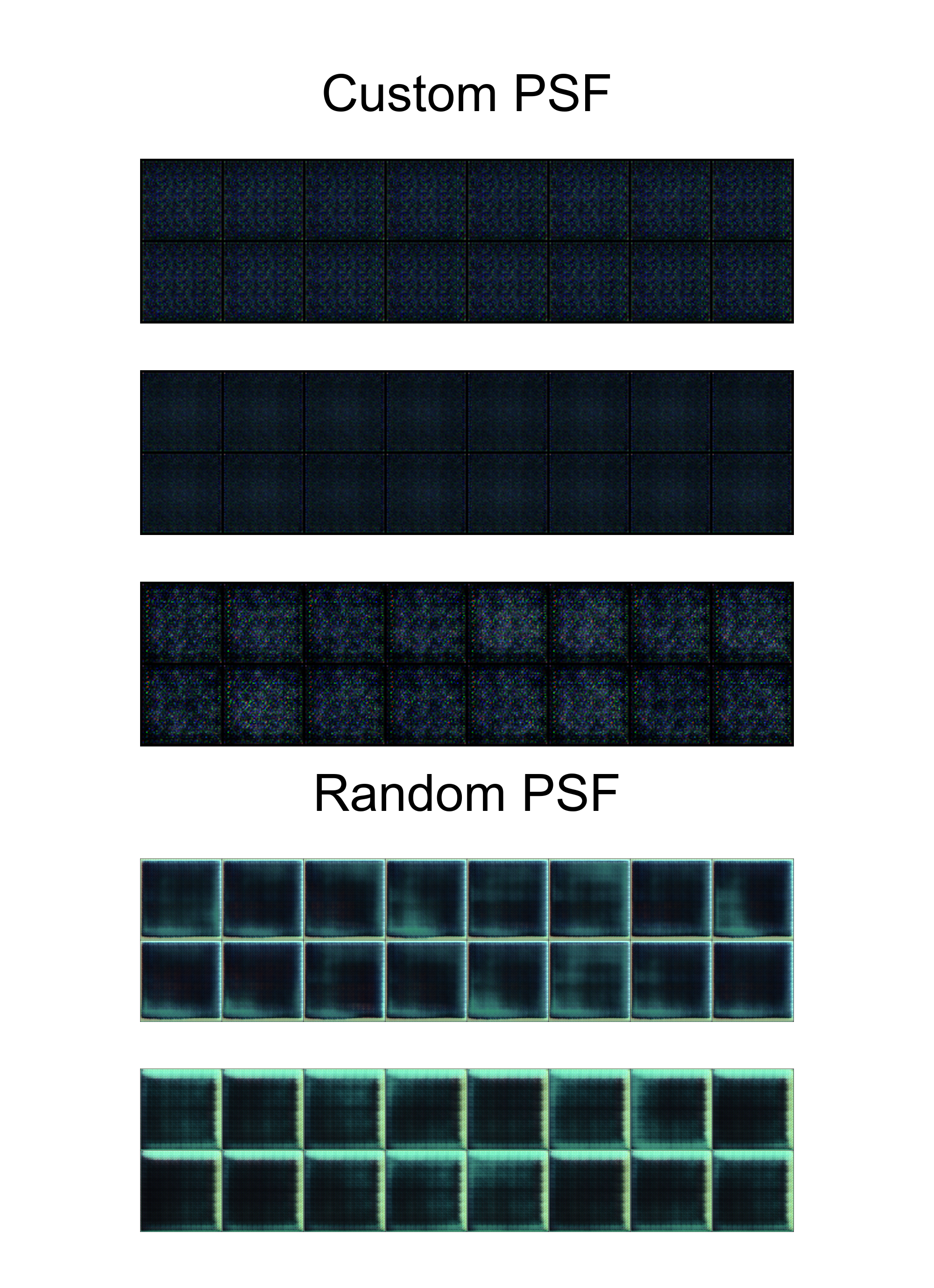}
\caption{
Results from our previous PISGAN baseline using different PSF modeling strategies.  
Top three rows: physically-motivated \textbf{Custom PSF}.  
Bottom two rows: \textbf{Random PSF}.  
While PISGAN already outperforms conventional baselines, our final STRCGAN (Fig.~\ref{fig:strcgan_main}) shows significantly superior detail preservation and artifact suppression.
}
\label{fig:pisgan_ablation}
\end{figure}

\end{appendix}

\end{document}